\documentclass[sigconf, nonacm]{acmart}
\AtBeginDocument{%
  }

\copyrightyear{2026}
\acmYear{2026}
\setcopyright{cc}
\acmConference[CF '26]{Proceedings of the 23rd ACM International Conference on Computing Frontiers}{May 19--21, 2026}{Catania, Italy}
\acmBooktitle{Proceedings of the 23rd ACM International Conference on Computing Frontiers (CF '26), May 19--21, 2026, Catania, Italy}
\acmDOI{10.1145/3801487.3806064}
\acmISBN{979-8-4007-2568-5/2026/05}
\usepackage{multirow}
\begin{document}

\newcommand{\nomemodello}{SeT-Diff}

\title{\nomemodello: Towards Semantic Foundation Models for HPC Telemetry and Time-Series}

%
\author{Giovanni B. Esposito}
\email{g.esposito@unibo.it}
\affiliation{%
  \institution{University of Bologna}
  \city{Bologna}
  \country{Italy}
}

\author{Francesco Antici}
\email{francesco.antici@unibo.it}
\affiliation{%
  \institution{University of Bologna}
  \city{Bologna}
  \country{Italy}
}

\author{Daniele Cesarini}
\email{d.cesarini@cineca.it}
\affiliation{%
  \institution{CINECA}
  \city{Bologna}
  \country{Italy}
}

\author{Andrea Bartolini}
\email{a.bartolini@unibo.it}
\affiliation{%
  \institution{University of Bologna}
  \city{Bologna}
  \country{Italy}
}


\begin{abstract}
Data centers and their compute nodes require accurate and flexible digital twins capable of modeling the complex interplay of workloads, environmental parameters, and physical metrics. Current machine learning approaches for HPC and its telemetry typically rely on a static subset of anonymous, fixed-position sensor variables tailored to single tasks. Consequently, these models become obsolete when target tasks change or sensor metrics vary. We propose \nomemodello{}, the first foundational model for compute node telemetry and time-series. Unlike rigid architectures, our diffusion-based approach conditions the generative process on each sensor's semantic description, decoupling the system dynamics from the structure of the dataset. Experiments on a real-world supercomputer dataset demonstrate a Mean Absolute Error (MAE) of 0.0470 on reconstruction tasks. \nomemodello{} exhibits zero-shot permutation stability, maintaining accuracy with negligible degradation even when sensors are shuffled. A single pre-trained model effectively performs data imputation, forecasting, and virtual sensing - achieving a 0.033 MAE in thermal inference - making \nomemodello{} an effective data-driven digital twin for HPC systems.
\end{abstract}

\ccsdesc[500]{Computing methodologies~Model development and analysis}
\ccsdesc[300]{Computing methodologies~Artificial intelligence}
\ccsdesc[300]{Computing methodologies~Multi-task learning}

\keywords{HPC Telemetry, Diffusion Models, Semantic Conditioning, Foundation Models, Virtual Sensing, Zero-Shot Adaptation}

 \maketitle

\section{Introduction}
High-performance computing (HPC) systems generate continuous streams of multivariate telemetry data, instrumental for system management \cite{borghesi2023m100, antici2025f}, predictive maintenance \cite{lima2021smart}, anomaly detection \cite{molan2023ruad}, and digital twins \cite{maiterth2024,maiterth2025hpc}. However, modeling this data is challenging due to high dimensionality, missing values \cite{medaiyese2025hardware}, and dynamically changing sensor layouts. Traditional time series models treat input signals as anonymous, fixed-position numerical vectors, causing them to fail when configurations vary.

To address this semantic gap, we propose \nomemodello{}, a context-aware foundation model based on the denoising diffusion framework. \nomemodello{} conditions the generation of multivariate time-series not only on historical data but also on the semantic context of the sensors. By integrating pre-trained textual embeddings (via Sentence-BERT \cite{reimers2019sentencebert}) describing each feature, the model associates a signal's physical behavior with its meaning rather than its positional index.

Our core contribution is a unified framework capable of zero-shot generalization across multiple downstream tasks—including data imputation, system load forecasting, and unmeasured metric regression (virtual sensing)—without architectural changes. Evaluated on the Marconi100 supercomputer dataset \cite{borghesi2023m100}, \nomemodello{} achieves a MAE of 0.0470 in reconstruction tasks. Furthermore, it maintains this accuracy indistinguishably between fixed and randomly shuffled sensor layouts (OLS slope of 0.99) and achieves an MAE of 0.033 in zero-shot thermal regression. This establishes \nomemodello{} as a versatile and robust digital twin for HPC systems.

\section{Background and Related Work}
\label{sec:back}
As HPC systems scale to support generative Artificial Intelligence (AI) and gigawatt-scale data centers, the complexity of their monitoring infrastructure has grown exponentially. Modern Tier-0 systems integrate diverse accelerators, multi-tier memory hierarchies, and liquid cooling, generating gigabytes of high-frequency telemetry daily \cite{borghesi2023m100,jadon2021challenges}. However, the usage of this data is currently limited to dashboarding and visualization.

AI adoption in system monitoring has been fragmented into specialized models tailored to specific tasks, such as autoencoders for anomaly detection \cite{molan2023ruad} or graph networks for thermal regression \cite{guindani2024exploring}. These supervised and unsupervised models suffer from a "semantic gap": they map anonymous numerical vectors based strictly on positional indices. If the monitored metrics change order or composition, these rigid pipelines become obsolete. Futhermore, these approaches are single task and trained on specific output features which cannot be changed at inference time.

To move beyond single-task models, recent works on Time Series Foundation Models, such as Time-LLM \cite{jin2024time} and Chronos \cite{ansarichronos}, adapt Transformer backbones for broad generalization. However, they function as deterministic mappings lacking the unified generative capabilities required for complex multi-task scenarios. Conversely, Denoising Diffusion Probabilistic Models \cite{ho2020denoising} provide a powerful, non-autoregressive framework for synthesizing high-fidelity data and handling unobserved values as noise for simultaneous forecasting and imputation. Yet, existing diffusion approaches for time-series (e.g., CSDI \cite{tashiro2021csdi}) remain structurally rigid and semantically blind. \nomemodello{} bridges this gap by fusing the generative versatility of diffusion models with the semantic awareness of language models, enabling true zero-shot adaptation to evolving hardware topologies.

\section{\nomemodello{} Foundational Model}
\label{sec:model}
We propose \nomemodello{}, the first multi-task foundational model for compute node telemetry, acting as a flexible digital twin. The core design philosophy decouples the modeling of physical system dynamics from the rigid structural constraints of sensor instrumentation. Traditional models approximate the conditional distribution $p(x_{t+1}|x_{0:t})$ assuming a fixed index $i$ for each feature. In contrast, \nomemodello{} treats the multivariate time series not as a fixed matrix, but as a collection of interacting physical signals defined by semantic identity and statistical behavior.

We leverage a DDPM framework for its ability to model complex distributions and its inherent robustness to noise. The training objective reverses a forward diffusion process adding Gaussian noise to the input data $X \in \mathbb{R}^{N \times P}$. Unlike standard approaches, our denoising network $\epsilon_\theta$ is explicitly conditioned on a rich context $\mathcal{C}$ describing what each sensor measures. This allows learning permutation-invariant representations: if "CPU Temperature" moves channels, the model recognizes its context embedding rather than its position, ensuring consistent generation across reconfigured hardware layouts.

\subsection{Context Encoding}
To transform raw metadata into a machine-interpretable signal, a dual-stream Context Encoder generates a context vector $\mathcal{C}$ for each of the $P$ features, composed of two embedding types:

\paragraph{Semantic Embeddings ($E_{text}$)} The primary anchor for permutation stability is the textual description of each sensor (e.g., "Temperature in CPU Core 0, in celsius degrees"). These define only the feature's identity, explicitly excluding behavioral characteristics or inter-feature relationships, which the model learns autonomously. We utilize Sentence-BERT to encode these into fixed-size dense vectors $d \in \mathbb{R}^{P \times 384}$. These embeddings provide semantic meaning, freeing the model from positional reliance and enabling zero-shot adaptation when sensors are reordered, removed, or newly introduced.

\paragraph{Statistical Descriptors ($E_{stat}$)} Semantic embeddings lack the scale or stationary properties crucial for normalizing the generative process. We compute a summary statistics vector $s \in \mathbb{R}^{P \times 9}$ for each channel (mean, quantiles, skewness, kurtosis). These descriptors provide a structural prior about the expected distribution, stabilizing generation for signals with vastly different dynamic ranges (e.g., RPM vs. Volts).

The final context $\mathcal{C}$ is obtained by projecting and concatenating these components, creating a conditioning tensor that accompanies the noisy input throughout denoising.

\subsection{Architecture}
The backbone of \nomemodello{} is a specialized Transformer architecture handling both temporal dependencies (dynamics) and inter-sensor correlations (system topology).

\paragraph{Input Projection} The noisy input $X_t$ and context $\mathcal{C}$ are projected into a common hidden dimension $H$. To seamlessly integrate the context, we prepend the embeddings to the raw time-series. Specifically, statistical and textual embeddings act as two additional "timesteps" per sensor, resulting in an augmented tensor $Z_t \in \mathbb{R}^{(N+2) \times P \times H}$. The attention mechanism thus attends to the sensor's description and statistics exactly as it attends to its past values.

\paragraph{Factorized Attention Mechanism} To efficiently model dependencies in high-dimensional telemetry ($P \gg 100$), we employ a factorized attention scheme alternating between two blocks:
\begin{itemize}
    \item \textbf{Temporal Attention:} Computes self-attention across the $N$ timesteps independently for each feature, capturing individual temporal evolution.
    \item \textbf{Feature-wise Attention:} Computes self-attention across the $P$ features independently per timestep. This captures instantaneous correlations (e.g., power impact on temperature), learning the system's "interaction graph."
\end{itemize}
This modular design is critical for handling missing data. By masking within attention heads, SeT-Diff ignores missing features or timesteps, ensuring generation is based solely on observed evidence. Crucially, appropriately modifying these masking bits during generation programs different tasks in \nomemodello{}.

\subsection{Training Objective}
Following the DDPM framework, our Transformer backbone $f_\theta(Z_t, t)$ estimates the noise added to the original data $X_0$ at time $t$. The training loss $\mathcal{L}_\theta$ is the distance between the estimated noise $\hat{\epsilon}$ and actual noise $\epsilon$:
$$ \mathcal{L}_\theta = \mathbb{E}\left[||\epsilon_t - f_\theta(Z_t, t)||_2^2\right] $$

\subsection{Generation and Multi-Task Inference}
The inference process is iterative and stochastic. Starting from pure Gaussian noise $X_T \sim \mathcal{N}(0, I)$, the model progressively estimates the noise to remove to reach $X_0$:
$$ X_{t-1} = \frac{1}{\sqrt{\alpha_t}} \left( X_t - \frac{1 - \alpha_t}{\sqrt{1 - \bar{\alpha}_t}} \hat{\epsilon}_\theta(Z_t, t) \right) + \sigma_t z $$


\section{Multi-task Capabilities and Inference Strategies}
Unlike specialized architectures, \nomemodello{} serves as a unified framework requiring no retraining. By leveraging the generative reverse diffusion process, we address distinct telemetry tasks solely by modifying a binary conditioning mask $M$ at inference time. Formally, the inference process iteratively samples missing parts $X_{miss}$ conditioned on observed data $X_{obs}$ and the semantic context $\mathcal{C}$.

\subsection{Imputation and Data Recovery}
Telemetry streams often suffer from missing data points due to transient sensor or network failures. We frame imputation as a random in-painting task. The mask $M_{imp}$ reflects the arbitrary pattern of missing values. During reverse diffusion, the model harmonizes unobserved entries with observed ones via learned inter-feature correlations (e.g., reconstructing "CPU Temperature" using "Fan Speed" and "Power Load"), ensuring coherent log reconstruction under severe data loss.

\subsection{Virtual Sensing (Regression)}
When comprehensive instrumentation is prohibitively expensive, operators infer "hidden" physical quantities from available proxies. We approach this as a feature-wise regression. To infer an unmonitored metric $j$, mask $M_{reg}$ zeros out the $j$-th column for all timesteps ($M_{:,j} = 0$), while proxy columns remain visible. Lacking numerical history for feature $j$, the model relies entirely on its Semantic Context (e.g., "DRAM Power") to understand what physical quantity to generate, inferring values based on the dynamics of visible proxies.

\subsection{Probabilistic Forecasting}
Deterministic forecasters often fail to capture the stochastic nature of workloads for proactive system management. We treat forecasting as a temporal extension task. Given a history context length $L$ and forecast horizon $H$, mask $M_{fore}$ observes timestamps $t \in [1, L]$ and masks $t \in [L+1, L+H]$. By running stochastic denoising multiple times, SeT-Diff generates a distribution of possible futures. This ensemble of trajectories inherently quantifies aleatoric uncertainty, providing confidence intervals for critical metrics like peak temperature.

\begin{figure*}
    \centering
    \includegraphics[width=\linewidth]{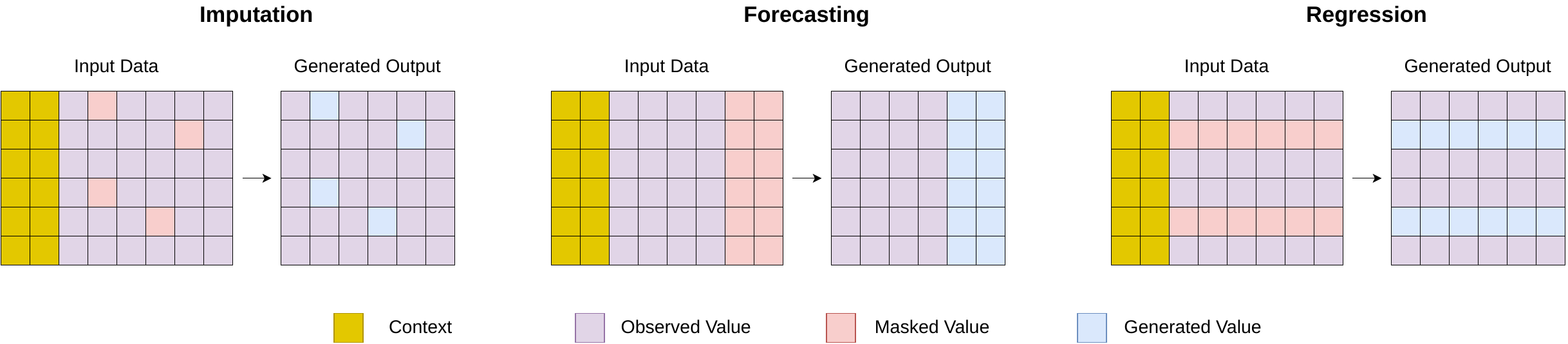}
    \caption{Unified Task Inference via Dynamic Masking Strategy. The diagram illustrates how a single pre-trained \nomemodello{} addresses distinct challenges by modifying the input mask during inference: (a) Imputation of random missing values; (b) Probabilistic Forecasting of future timesteps; (c) Virtual Sensing (Regression) of entire unobserved features}    
    \label{fig:masking}
\end{figure*}

\section{Experimental Evaluation}
\label{sec:eval}
We validate \nomemodello{} on real-world HPC telemetry, focusing on multi-task capability, zero-shot permutation stability, and the efficacy of semantic conditioning.

\subsection{Experimental Setup}
\paragraph{Dataset \& Preprocessing}
We use the M100 ExaData dataset \cite{borghesi2023m100} (20 months of Marconi100 operations), aggregating out-of-band (IPMI) and in-band (Ganglia) metrics. Data is structured into rolling windows ($L=32$), yielding $\sim$40K windows with $P=261$ metrics. We applied chronological splitting to prevent look-ahead bias and standardized features using training-set statistics Missing values in training use mean imputation; test-set gaps remain intact to simulate real sparsit

\paragraph{Context \& Implementation}
Sensor descriptions were encoded via Sentence-BERT into fixed embeddings $E_{text} \in \mathbb{R}^{261 \times 384}$. For production feasibility, we employed a compact Transformer (1 block, $H=64$, 8 heads, 136.3K parameters) with $T=1000$ diffusion steps. Training on a single NVIDIA A100 GPU (Adam, LR=$10^{-3}$, batch 128) took $\sim$90 minutes for 50 epochs. Generating a $32 \times 261$ window takes $\sim$140 ms, proving lightweight efficiency.

\subsection{Results}
Table \ref{tab:main_results} reports the performance across operational scenarios under stochastic test-time perturbations. We compare \nomemodello{} against a structurally identical diffusion baseline that relies on standard positional indices instead of textual semantic embeddings. While Mean Absolute Error (MAE) serves as our primary comparative metric, we additionally conduct an extended probabilistic evaluation on our proposed model using the Continuous Ranked Probability Score (CRPS) to quantify its generative uncertainty.

\begin{table}[h]
    \centering
    \caption{Performance comparison. \nomemodello{} outperforms the structurally identical positional baseline in point-wise accuracy (MAE). Additionally, we report the CRPS for our proposed model to highlight its well-calibrated probabilistic uncertainty quantification.}
    \label{tab:main_results}
    \resizebox{\columnwidth}{!}{%
    \begin{tabular}{llccc}
    \toprule
    \multirow{2}{*}{\textbf{Operational Task}} & \multirow{2}{*}{\textbf{Perturbation}} & \textbf{Baseline} & \multicolumn{2}{c}{\textbf{\nomemodello{}}} \\
    \cmidrule(lr){3-3} \cmidrule(lr){4-5}
    & & \textbf{MAE} & \textbf{MAE} & \textbf{CRPS} \\
    \midrule
    \textit{Imputation} & None (Clean Data) & 0.154 & \textbf{0.0470} & \textbf{0.0299} \\
    \textit{Imputation} & Shuffled Features & 0.154 & \textbf{0.0472} & \textbf{0.0298} \\
    \textit{Virtual Sensing} & Masked Features ($\leq$ 50\%) & 0.156 & \textbf{0.0622} & \textbf{0.0589} \\
    \textit{Forecasting} & Masked Future (8 timesteps)  & 0.155 & \textbf{0.0613} & \textbf{0.0583} \\
    \bottomrule
    \end{tabular}%
    }
\end{table}
\paragraph{Impact of Semantic Conditioning}
The integration of semantic context yields massive performance gains over the positional baseline. In standard imputation, semantic embeddings significantly reduces the error from a 0.154 MAE to a 0.047 MAE. This superiority is pronounced also under virtual sensing and forecasting tasks, showing how \nomemodello{} effectively leverages the remaining contextual descriptions to guide the generation, bounding the degradation to a MAE of $\sim0.062$.

\paragraph{Operational Performance and Zero-Shot Stability}
Beyond outperforming the baseline, \nomemodello{} demonstrates remarkable versatility as a unified modeling tool. Under severe masking (up to 50\% feature loss), it acts as a robust soft-sensor (0.0622 MAE, 0.0589 CRPS) by capturing physical inter-dependencies. Similarly, it provides reliable temporal extrapolations for forecasting (0.0613 MAE, 0.0583 CRPS). 
Under random permutation of the input sensor array , \nomemodello{} yields performance virtually identical to its unperturbed state (0.0472 MAE vs. 0.0470 MAE). This confirms that the model identifies signals strictly by their semantic meaning, exhibiting intrinsic zero-shot robustness to hardware reconfigurations where traditional (index-based) models naturally fail. Furthermore, the CRPS values obtained for our model (e.g., 0.0299 for imputation) are consistently lower than the corresponding MAE. This confirms that the model's generated predictive distributions are also sharp and well-calibrated, providing operators with highly reliable confidence intervals

\subsection{Operational Use-Cases (Virtual Sensing)}
We defined three Virtual Sensing scenarios (Table \ref{tab:case_studies}), masking the entire timeseries if selected target features to evaluate \nomemodello{}'s reconstructive capabilities using solely the context and the remaining sensors sequences.

\begin{table}[h]
    \centering
    \caption{Virtual Sensing Case Studies: Input contexts vs. Targets. In brackets the \# of features.}
    \label{tab:case_studies}
    \resizebox{\columnwidth}{!}{%
    \begin{tabular}{lllc}
    \toprule
    \textbf{Scenario} & \textbf{Input Context} & \textbf{Target} & \textbf{MAE} \\
    \midrule
    \textit{A. Power Est.} & CPU+GPU Load (158) & Node Powers (22) & \textbf{0.0880} \\
    \textit{B. Thermal Est.} & Power+Ambient (23) & Core Temps (66) & \textbf{0.0332} \\
    \textit{C. GPU Est.} & Sec. GPU Metrics (124) & GPU Pwr/Temp (28) & \textbf{0.154} / \textbf{0.025} \\
    \bottomrule
    \end{tabular}%
    }
\end{table}

\begin{itemize}
    \item \textbf{A. Power Estimation:} \nomemodello{} can effectively map computational activity (OS load) to power consumption without requiring physical instrumentation, achieving a 0.088 MAE.
    \item \textbf{B. Thermal Estimation:} The model can identify the lagged correlation between power injection and temperature rise, to reconstruct internal thermal maps purely from power/ambient context, with a 0.033 MAE
    \item \textbf{C. GPU Estimation:} Leveraging microarchitectural proxies (clocks, utilization), the model is able to impute GPU temperatures (0.0250 MAE), though power recovery proves slightly more complex (0.1545 MAE), an aspect reserved for future investigation.
\end{itemize}

\section{Conclusion and Future Work}
\label{sec:conclusion}

We presented \nomemodello{}, a context-aware diffusion framework advancing the realization of Foundation Models for HPC telemetry. By integrating semantic knowledge directly into the generative process, we address a pervasive challenge in data center operations: the rigidity of data-driven models in the face of dynamic hardware configurations.

Experiments on the M100 Exascale dataset demonstrate that \nomemodello{} is a robust system-level tool. Conditioning on textual embeddings grants it unique "permutation stability," maintaining consistent performance even when sensor layouts are completely reshuffled—a scenario that traditional models can't handle. Furthermore, a single pre-trained model seamlessly handles data imputation, virtual sensing, and forecasting simply by altering the inference-time masking strategy. This "train once, deploy everywhere" capability significantly reduces operational overhead by eliminating the need to train specialized models for every compute node variant.

Future work will scale this paradigm along two axes. First, we will expand the semantic vocabulary to include system topologies and executing processes. Second, we will extend the training corpus across multiple architectures and heterogeneous sources.
\begin{acks}
The activities of this work have been supported by EU - NextGenerationEU 
with funds made available by the National Recovery and Resilience Plan (PNRR) Mission
4, Component 2, Investment 3.3 (D.M. 117/2023), EuroHPC EUPEX (g.a. 101033975), EuroHPC JU SEANERGYS (g.a. 101177590), and CINECA
\end{acks}

\bibliographystyle{ACM-Reference-Format}
\bibliography{bib-short}

\appendix

\end{document}